\documentclass[conference]{IEEEtran}
\IEEEoverridecommandlockouts
\usepackage{cite}
\usepackage{amsmath,amssymb,amsfonts}
\usepackage{algorithmic}
\usepackage{graphicx}
\usepackage{textcomp}
\usepackage{xcolor}
\def\BibTeX{{\rm B\kern-.05em{\sc i\kern-.025em b}\kern-.08em
    T\kern-.1667em\lower.7ex\hbox{E}\kern-.125emX}}

\usepackage{booktabs} 
\usepackage{paralist}
\usepackage{enumitem}
\usepackage{url}
\usepackage{authblk}
\usepackage{paralist}

\renewcommand{\vec}[1]{\mathbf{#1}}

\newcommand{\pow}{\mathcal{P}}

\let\emptyset\varnothing

\begin{document}

\title{Immigration Document Classification and Automated Response Generation \\
}

\author[1]{Sourav Mukherjee}
\author[2]{Tim Oates}
\author[3]{Vince DiMascio}
\author[2]{Huguens Jean}
\author[4]{Rob Ares}
\author[3]{David Widmark}
\author[3]{Jaclyn Harder}
\affil[1]{Fairleigh Dickinson University, Vancouver, Canada. Email: sourav.mukherjee.8@gmail.com}
\affil[2]{Synaptiq, USA. Email: \{tim.oates, huguens.jean\}@synaptiq.ai}
\affil[3]{Berry Appleman \& Leiden LLP, USA. Email: \{Vdimascio, dwidmark, jaharder\}@balglobal.com}
\affil[4]{Independent, USA. Email: rob@hireredrectangle.com}

\maketitle

\begin{abstract}

In this paper, we consider the problem of organizing supporting documents vital to U.S.~work visa petitions, as well as responding to Requests For Evidence (RFE) issued by the U.S.~Citizenship and Immigration Services (USCIS). Typically, both processes require a significant amount of repetitive manual effort. To reduce the burden of mechanical work,  we apply machine learning methods to automate these processes, with humans in the loop to review and edit output for submission. In particular, we use an ensemble of image and text classifiers to categorize supporting documents. We also use a text classifier to automatically identify the types of evidence being requested in an RFE, and used the identified types in conjunction with response templates and extracted fields to assemble draft responses.  Empirical results suggest that our approach achieves considerable accuracy while significantly reducing processing time.
\end{abstract}

\section{Introduction}
\label{sec:Introduction}

Preparing a U.S.~work visa (e.g., H-1B, TN) petition requires compiling a large number of supporting documents, e.g., passport and visa pages, driver's license, I-797 \cite{I797}, I-797C \cite{I797C}, Employment Authorization Document \cite{EAD}, certificates, transcripts, and so on. Any immigration law firm retained by a large corporation may need to simultaneously pursue hundreds of such applications, resulting in a plethora of documents that must be systematically stored and handled. While document categorization is a necessary first step, relying entirely on humans for this does not scale  with data volume. Moreover, categorization of the above document types is generally a mechanical process that does not require ingenuity, and therefore is a good candidate for automation. However, these documents are usually provided by beneficiaries of the petitions as scanned files and facsimiles of varying image quality, thereby making optical character recognition (OCR) difficult. Consequently, automatic classification of these documents based on textual content alone is error prone.  We argue that classification can be made more accurate by relying on content as well as appearance. Therefore, we adopt a novel approach of employing an ensemble of classifiers, one based on textual content, and another based on visual content, for classifying supporting documents.

Once a petition has been submitted, the office of U.S.~Citizenship and Immigration Services (USCIS) frequently requests additional supporting information by sending a Request For Evidence (RFE) to the petitioner. An RFE may be issued based on any one or more of the reasons listed by USCIS in \cite{RFE}. E.g., USCIS may ask for more clinching evidence that the beneficiary is seeking to work in a \textit{specialty occupation}, or that the beneficiary is qualified for the position, among others.  In this paper, we refer to every such reason as an RFE attack. To successfully respond to the RFE, the law firm preparing the petition reviews the RFE to identify every RFE attack contained therein. Thereafter, the firm compiles a response that addresses every attack with supporting information and documents. Typically, this process is entirely manual and time consuming. However, RFEs with the same types of attack often have similar language. E.g., RFEs for two beneficiaries seeking to work as Computer Systems Analysts (Specialty Occupation Code, SOC : 15-1211; see \cite{SOC}, page 18), may both ask for proof that the beneficiary has a baccalaureate or more advanced degree, that such a degree is necessary for performing the job, that the employer typically requires this qualification for the position, and so on. Unsurprisingly, the responses to these RFEs also tend to have similar language. Therefore, response templates are often used to avoid having to author each response from scratch. Once all RFE attacks have been identified, the next steps consist of selecting response templates, inserting beneficiary-specific information into the templates' placeholders, and compiling the filled templates into a draft response ready to be reviewed by a lawyer, who then modifies the content based on domain expertise prior to sending the response to USCIS. 
We explore whether this workflow is suitable for partial automation, not to obviate human experts, but rather, to automate rote work, thereby freeing up more time for them to fine tune responses strategically.
Since RFEs originate from a single source, namely, USCIS, variability in scan quality is lower compared to those of supporting documents discussed in the previous paragraph, making RFEs more amenable to text extraction through OCR. We demonstrate that automatic identification of attack types within an RFE is indeed possible by training classifiers based on textual content. We further show that the identified attack types, together with data extracted from the RFE and queried from a beneficiary database, may be used to identify appropriate response templates and populate the templates' placeholders, resulting in a draft response ready for human review. Empirical results indicate that the above automation reduces manual effort and improves turnaround time substantially while achieving high accuracy.

The rest of this paper is organized as follows. Section \ref{sec:RelatedWork} reviews related work while Section \ref{sec:Methodology} describes our novel contributions. Section \ref{sec:Evaluation} provides empirical evidence in support of our approach while \ref{sec:Discussion} interprets the results. Finally, Section \ref{sec:Conclusion} concludes this paper.

\section{Related Work}
\label{sec:RelatedWork}

Applications of machine learning techniques to problems in the legal domain have become increasingly popular in recent years. Specific areas where machine learning has been applied include outcome prediction, e-discovery, document categorization, contract review/due diligence, automated document assembly, information retrieval, document translation, legal analytics, and so on. In this section, we briefly review research relevant to our work. For more comprehensive surveys, please refer to \cite{Surden2014,10.1007/978-3-030-19823-7_31,Faggella2020}.

\subsection{Outcome Prediction}
\label{subsec:OutcomePrediction}

The Supreme Court Forecasting Project \cite{10.2307/4099370} used classification trees with six input parameters, namely, circuit of origin, issue area, petitioner type, respondent type, whether the lower court ruling was pro-liberal or pro-conservative, and whether the unconstitutionality of a practice was included in the petitioner's argument, to predict U.S.~Supreme Court outcomes. Another classification tree approach to the same problem was reported in \cite{10.2307/3688543}. In \cite{Katz2016}, random forest classifiers were used to predict U.S.~Supreme Court outcomes over a much broader time window, namely, from 1816 to 2015. Similar predictive models have been explored in other legal jurisdictions as well. For example, support vector machine (SVM) classifiers were trained on data from 
European Court of Human Rights cases to predict violation of human rights convention articles; the training data in these cases was textual, featurized using $n$-grams and topics (word clusters) in \cite{Aletras2016PredictingJD}, and using TF-IDF vectorization on $n$-grams $(n \leq 4)$ in \cite{Medvedeva2019UsingML}.

\subsection{E-Discovery}
\label{subsec:EDiscovery}

Given a legal matter and a large collection of documents, e-discovery refers to the process of identifying those documents that are most relevant to the matter, and filtering out irrelevant documents. This process is also sometimes referred to as \textit{predictive coding} or \textit{technology assisted review} \cite{Surden2014}. Viewing e-discovery as a binary classification problem allows the application of standard supervised learning algorithms to address it. In \cite{DBLP:conf/icail/YangGFY17}, the performance of several such algorithms, namely, support vector machine (SVM), logistic regression, XGBoost, multi-layered perceptron (MLP), and 1-nearest neighbor, have been compared on a standard e-discovery benchmark. One of the challenges in e-discovery is the shortage of \textit{labeled} data. While the volume of electronically stored documents have grown by orders of magnitude, it is unrealistic to expect the number of human labeled examples to grow at the same rate. As a result, iterative training \textit{protocols} have been used that begin with a labeled seed dataset (obtained using a keyword search or random selection) for training a classifier, and then gradually augment the labeled training dataset and retrain. Three well known protocols, namely, Continuous Active Learning (CAL), Simple Active Learning (SAL), and Simple Passive Learning (SPL) are compared in \cite{Cormack2014EvaluationOM}.

\subsection{Document Categorization}
\label{subsec:DocumentCategorization}

Even after filtering out irrelevant documents through e-discovery, the number of documents that need to be reviewed for a case may be in the hundreds. Therefore, organizing large sets of documents into manageable categories is essential. To address this problem, both supervised approaches that assume document categories to be known \textit{a priori} \cite{Lemley2007,DBLP:conf/bigdataconf/WeiQYZ18,Silva2018DocumentTC,DBLP:conf/fedcsis/UndaviaMO18}, and unsupervised approaches where documents are clustered based on similarity without any prior knowledge of categories \cite{DBLP:conf/cikm/LuCAK11,DBLP:conf/propor/FurquimL12,Kumar2012}, have been proposed.

\subsection{Legal Drafting}
\label{subsec:LegalDrafting}

Since legal documents of the same kind often tend to use similar language, law firms frequently use templates with placeholders for drafting new documents by populating the templates' placeholders with appropriate values. For example, contracts between service providers and customers/clients often tend to include similar clauses. Unsurprisingly, automated legal drafting, also known as automated document assembly, has been an active area of research for at least the past three decades \cite{DBLP:conf/afips/SprowlBCEK84}. As of this writing, machine learning based approaches to drafting are viewed as promising \cite{Betts2017,Miller}.

To the best of our knowledge, there is no published work on machine learning approaches to documents related to immigration petitions. Unlike existing document categorization approaches, we adopt an ensemble approach for classifying supporting documents based on content (i.e., text classification) as well as appearance (i.e., image classification). While existing approaches to legal outcome prediction use classifiers that predict outcomes based on featurized (i.e., vectorized) text, we use text similarity between the RFE document and known text fragments to predict the presence of an attack type. Unlike existing legal drafting approaches, our goal is to draft responses to RFEs as opposed to legal contracts.

\section{Methodology}
\label{sec:Methodology}
Formally, we define the following problems.

\subsection{Problem Statement}
\label{subsec:ProblemStatement}

\begin{itemize}
    \item \textbf{P1 (Supporting Document Classification).} Let $\mathcal{C}$ be a finite, known set of document classes, and dataset 
    $\mathbb{D}_1 = \{(x, y): x \mbox{ is a document, } y \in \mathcal{C} \mbox{ is its unique class}\}$. 
    Given any new document $x$, calculate the probability  $P(y|x)$ that the document belongs to class $y$ for every $y \in \mathcal{C}$.
    \item \textbf{P2 (RFE Attack Identification).} Let $\mathcal{A}$ be a finite, known set of RFE attack types, and dataset 
    $\mathbb{D}_2 = \{(x, y): x \mbox{ is a Request For Evidence (RFE), } y \in \pow(\mathcal{A}) - \emptyset \mbox{ is a nonempty set of attacks}\}$, where $\pow(.)$ denotes power set. Given a new RFE document $x$, predict the set of attacks $y$ contained therein.
    \item \textbf{P3 (RFE Response Generation).} Suppose the $B$ is the beneficiary of a petition, and $data(B)$ represents data about the beneficiary available to the petitioner. Suppose $y \in \pow(\mathcal{A}) - \emptyset$ is the set of RFE attacks identified in an RFE issued to the beneficiary. Generate a response draft based on $y$ and $data(B)$.
\end{itemize}

\subsection{Approach: Supporting Document Classification}
\label{subsec:SupportingDocumentClassification}

\begin{figure}[ht]
    \centering
    \includegraphics[scale = 0.40]{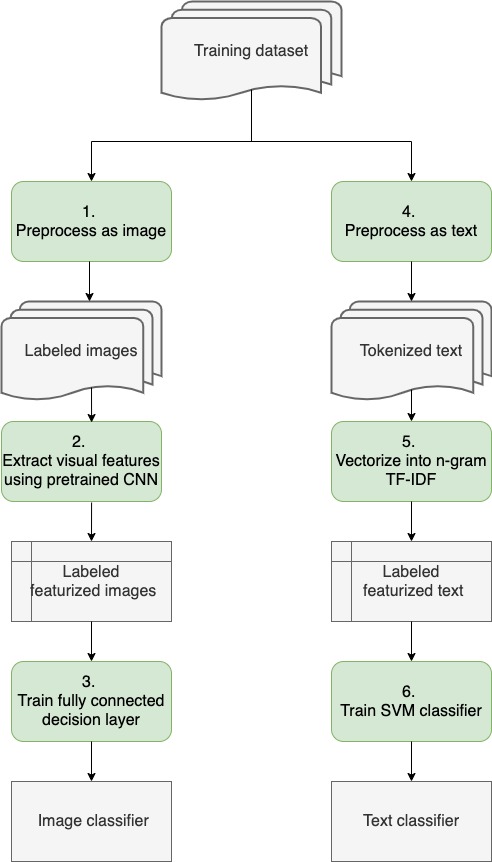}
    \caption{Training image classifier and text classifier based on labeled supporting documents.}
    \label{fig:ImageAndTextClassifierTraining}
\end{figure}

Recall from Section \ref{sec:Introduction} that the supporting documents for a work visa petition are usually received by the law firm from the beneficiary as scans or facsimiles of varying image quality. As a result, text extraction using optical character recognition (OCR) is error-prone, which may reduce the accuracy of a classifier based on textual content alone. To address this challenge, we use an ensemble of two classifiers, namely an image classifier that considers the appearance of a page, and a text classifier that considers its textual content. Figure \ref{fig:ImageAndTextClassifierTraining} depicts the process of training these classifiers based on the same training dataset.

\subsubsection{Training the Image Classifier}
\label{subsubsec:TrainingTheImageClassifier}

To train the image classifier, we convert each document in the training dataset into a set of images, one image per page, resulting in a dataset of labeled images, which are fed as training data to a convolutional neural network (CNN) classifier. Since training a CNN to learn visual features from scratch requires huge volumes of training data, CNN-based image classifiers frequently rely on transfer learning \cite{DBLP:conf/nips/YosinskiCBL14}, with lower level feature extractors pretrained on a different, much larger dataset, and higher level layers trained on the dataset of interest. Following this approach, we use the VGG-16 \cite{DBLP:journals/corr/SimonyanZ14a} architecture trained on the ImageNet dataset \cite{DBLP:conf/cvpr/DengDSLL009}, but excluding the final decision layer, as feature extractor. The weights of this CNN are frozen, i.e., never changed. We then append a fully connected decision layer to this CNN, and an output layer of size $|\mathcal{C}|$. The weights of this final decision layer are trained using labeled, featurized images, whereas the features themselves are extracted by the pretrained CNN.

\subsubsection{Training the Text Classifier}
\label{subsubsec:TrainingTheTextClassifier}

To train the text classifier, we first use optical character recognition (OCR) to extract text from the documents, and then tokenize the text by splitting by whitespace. Once a document has been tokenized, we represent it as an $n$-gram vector ($n \in \{2, 3\}$) weighted by term frequency-inverse document frequency (TF-IDF). Since Support Vector Machines (SVM) are known to achieve high accuracy in classifying text based on sparse vector representations \cite{DBLP:conf/ecml/Joachims98}, we use the resulting labeled feature vectors to train an SVM classifier.

\subsubsection{Predicting Document Type}
\label{subsubsec:PredictingDocumentType}

\begin{figure}[ht]
    \centering
    \includegraphics[scale = 0.45]{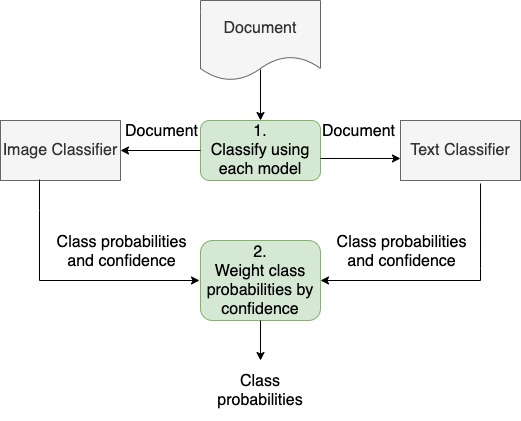}
    \caption{Prediction of document type by ensemble classifier.}
    \label{fig:ImageAndTextClassifierPrediction}
\end{figure}

Given a new document $x$, the ensemble model estimates the probability $P(y|x)$ that the document belongs to class $y$, for every $y \in \mathcal{C}$, as shown in Figure \ref{fig:ImageAndTextClassifierPrediction}.
Taking $x$ as input, the image classifier calculates $\{P_{\mathrm{image}}(y|x) : y \in \mathcal{C}\}$, i.e., its estimates of the document belonging to class $y$ for every $y \in \mathcal{C}$. The entropy of this distribution is given by
\begin{equation*}
    H_{\mathrm{image}}(x) = - \sum_{y \in \mathcal{C}} P_{\mathrm{image}}(y|x) \lg \left(P_{\mathrm{image}}(y|x)\right)
\end{equation*}
where $\lg(.)$ denotes logarithm to the base 2. 
Similarly, the text classifier  calculates its own estimates $\{P_{\mathrm{text}}(y|x) : y \in \mathcal{C}\}$ with entropy
\begin{equation*}
    H_{\mathrm{text}}(x) = - \sum_{y \in \mathcal{C}} P_{\mathrm{text}}(y|x) \lg \left(P_{\mathrm{text}}(y|x)\right)
\end{equation*}
We quantify the confidence $w$ of a classifier as the reciprocal of the above entropy. Thus,
\begin{equation}
    w_{\mathrm{image}}(x) = \frac{1}{\max(H_{\mathrm{image}}(x), \epsilon)}
\end{equation}
and
\begin{equation}
    w_{\mathrm{text}}(x) = \frac{1}{\max(H_{\mathrm{text}}(x), \epsilon)}
\end{equation}
where $\epsilon = 0.001$ is a small constant. To avoid divide-by-zero errors, $\max(H_{\mathrm{image}}(x), \epsilon)$ and $\max(H_{\mathrm{text}}(x), \epsilon)$ are used in the denominator instead of $H_{\mathrm{image}}(x)$ and $H_{\mathrm{text}}(x)$. Finally, the class probabilities estimated by the ensemble are calculated as follows:
\begin{equation}
    P(y|x) = \frac{w_{\mathrm{image}}(x)P_{\mathrm{image}}(y|x) + w_{\mathrm{text}}(x)P_{\mathrm{text}}(y|x)}
    {w_{\mathrm{image}}(x) + w_{\mathrm{text}}(x)}
\end{equation}

\subsection{Approach: RFE Attack Identification}
\label{subsec:RFEAttackIdentification}

Recall from Section \ref{sec:Introduction} that the office of U.S.~Citizenship and Immigration Services (USCIS), in response to a work visa petition,  frequently requests additional information by sending a Request For Evidence (RFE) to the petitioner, to establish more conclusively that, e.g., the beneficiary is seeking employment in a specialty occupation, has the necessary qualification, and so on. In this paper, we use the term \textit{RFE attack} to refer to reasons for issuing an RFE; the set of all possible reasons is listed in \cite{RFE}. The first step in producing a successful response is to identify all the RFE attacks contained within an RFE. In this section, we address this problem.  

\begin{figure}[ht]
    \centering
    \includegraphics[scale = 0.40]{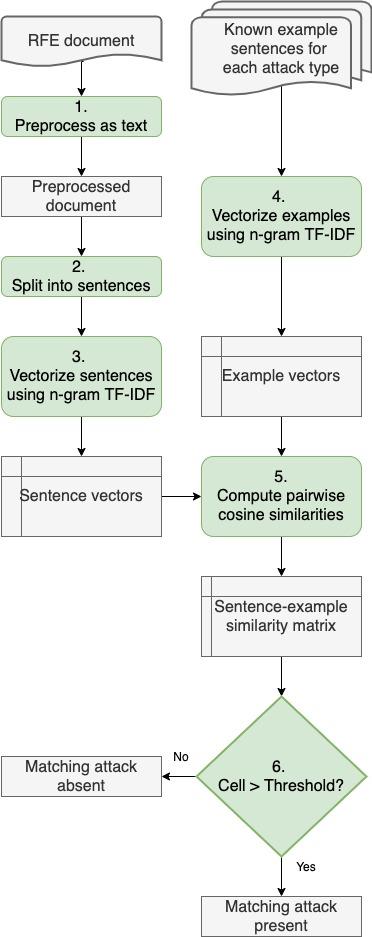}
    \caption{Workflow for identifying attack types within a Request For Evidence (RFE) document.}
    \label{fig:RFEAttackDetectionWorkflow}
\end{figure}

Since all RFEs originate from a single source, namely, USCIS, there is greater consistency in scanned image quality. Consequently, RFEs are more amenable to OCR based text extraction. We find  textual content thus extracted to be useful in attack type identification. On the other hand, the visual content, i.e., appearance, of an RFE does not vary across different attack types. Therefore, image classification is not a suitable approach for this problem. Figure \ref{fig:RFEAttackDetectionWorkflow} depicts the workflow for identifying RFE attacks contained in a document.
The starting point is an RFE document, $x$. The historical data used for predicting attack types consists of a set $\mathcal{E}$ of example sentences previously found in these types of attacks. The expectation is not that sentences identical to these will appear in the new RFE, but that semantically similar sentences may occur. Textual content, extracted using OCR, is preprocessed by converting text to lowercase, and removing stopwords, numeric information, and non-English characters. The preprocessed text is then split by one or more newlines yielding a set of sentences. After vectorizing both the RFE sentences and the example sentences to TF-IDF weighted $n$-gram representation ($n \in \{1, 2, 3\}$), we compute pairwise cosine similarities between every RFE sentence and every example sentence, resulting in a cosine similarity matrix. Finally, we predict that an attack is present in document $x$ if there is at least one sentence $s_1$ in the document and at least one sentence $s_2$ in the set of examples such that the cosine similarity between their vectorized representations is greater than a pre-defined threshold $\tau$ (we set $\tau = 0.6$ through manual tuning). For every such example $s_2$, we say that the attack represented by $s_2$ is present in the document. In other words, the decision rule is given by:
\begin{equation}
\begin{split}
    (\forall a \in \mathcal{A})[(\exists s_1 \in x, s_2 \in \mathcal{E})[(sim(\vec{s}_1, \vec{s}_2) > \tau)] \Rightarrow  \\ (attack(s_2) \in attacks(x)) ]
\end{split}
\end{equation}
where $\vec{s}_1, \vec{s}_2$ are TF-IDF weighted $n$-gram vector representations of sentences $s_1, s_2$, $sim(.,.)$ is the cosine similarity metric
\begin{equation*}
    sim(\vec{s}_1, \vec{s}_2) = \frac{\vec{s}_1 \cdot \vec{s}_2}{|\vec{s}_1||\vec{s}_2|}
\end{equation*}
$attack(s_2)$ is the attack type of example sentence $s_2$, and $attacks(x)$ is the set of attacks contained in the document $x$.
Identification of RFE attack types in a document makes automated drafting of the response possible, as we discuss next.

\subsection{Approach: RFE Response Drafting}
\label{subsec:RFEResponseDrafting}

\begin{figure}[ht]
    \centering
    \includegraphics[scale = 0.40]{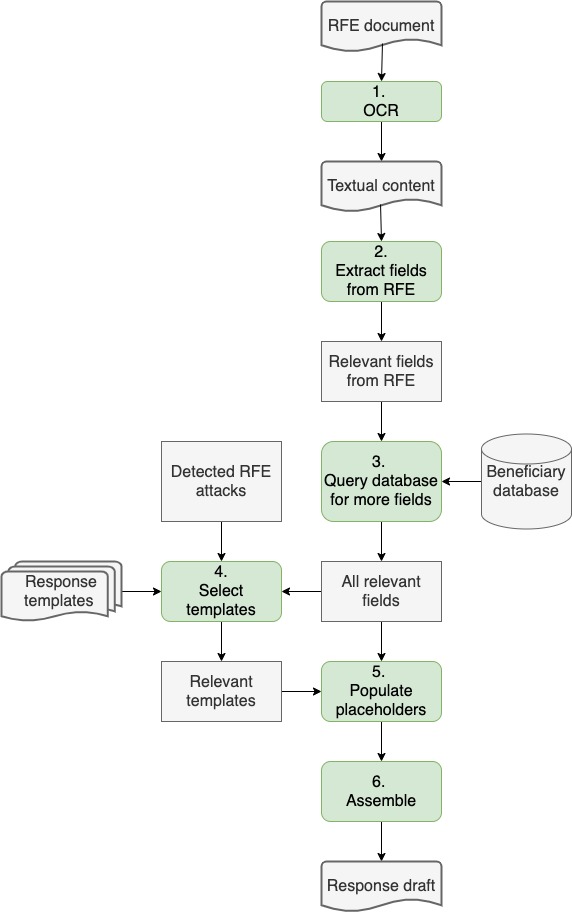}
    \caption{Workflow for preparing RFE response draft. This assumes that RFE attacks have already been identified using the workflow in Figure \ref{fig:RFEAttackDetectionWorkflow}.}
    \label{fig:RFEResponseWorkflow}
\end{figure}

Figure \ref{fig:RFEResponseWorkflow} depicts the workflow for drafting a response to an RFE, which assumes that the workflow for RFE attack type identification discussed in the previous section has already been completed.  
For preparing a response draft, we do not remove any stopwords or symbols from the text extracted using OCR, since these may be useful in extracting fields from the RFE. We find several essential fields to be readily extractable from RFEs using regular expression matching; these fields include case number, employee name, employer name, attorney name, date of the RFE, due date of the response, and so on. These fields are used to query a database containing additional data about the beneficiary, such as specialty occupation code (SOC) \cite{SOC}, field of study, degree received, name of institution, and so on. Extracting the above pieces of data from the RFE and the database is necessary for two reasons. First, determining which templates to select may depend on certain field values in addition to attack types; e.g., for attacks of type \textit{specialty occupation}, different specialty occupation codes \cite{SOC} require different response templates. Second, this data is also used to populate template placeholders. Once templates have been selected based on detected attack types and extracted fields, and placeholder values have been populated, these filled templates are concatenated and a preamble is added, resulting in a response draft ready for expert review. The next section presents empirical results.

\section{Evaluation}
\label{sec:Evaluation}

To empirically validate our approach, we use real-world end-to-end workflows used at a large law firm, and execute the workflows with and without the partial automation methods described in Section \ref{sec:Methodology}, using validation data that is completely disjoint from training data. Comparison of the total execution times with and without automation helps us quantify the advantage of our approach. We also report predictive accuracy scores. The following tools were used:
\begin{inparaenum}[(a)]
    \item programming language: Python 3,
    \item optical character recognition: Tesseract OCR \cite{10.5555/1288165.1288167},
    \item SVM and logistic regression: scikit-learn \cite{scikit-learn}, and
    \item CNN: Keras \cite{chollet2015keras} with Tensorflow \cite{abadi2016tensorflow}.
\end{inparaenum}

\subsection{Supporting Document Classification}
\label{subsec:Evaluation:SupportingDocumentClassification}

\subsubsection{Experimental Setup}

\begin{figure}[ht]
    \centering
    \includegraphics[scale = 0.35]{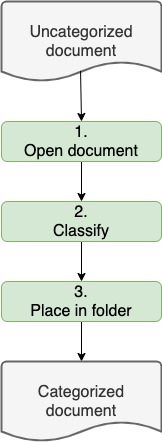}
    \caption{Experimental setup for supporting document type classification.}
    \label{fig:DocumentClassificationSetup}
\end{figure}

Figure \ref{fig:DocumentClassificationSetup} depicts the workflow for supporting document classification, and consists of opening an uncategorized document, classifying it, and placing the document in a folder determined by its category. In the approach being evaluated, each of the above steps is automated. In the baseline, each step is manual.

We use a dataset with 104 documents, where each document belongs to one of two types: I-797 approval, or I-797 receipt. We selected these categories to examine whether our classifier can distinguish between semantically different document types that have strong visual and textual similarities.

\subsubsection{Results}

\paragraph{Accuracy.} 
Prediction accuracy for the above dataset is shown in Table \ref{tbl:DocumentClassificationAccuracy}. Of the 104 documents used in the above evaluation, 102 were correctly classified, resulting in a prediction accuracy of 98.08\%. Of these 104 documents, all 33 out of 33 I-797 approvals were correctly classified, whereas 69 out of 71 I-797 receipts were correctly classified.
\begin{table}[ht]
    \centering
    \begin{tabular}{c c c c}
        \hline
         Document type & Count & Correct prediction count & Accuracy (\%) \\
         \hline
         All & 104 & 102 & 98.08 \% \\
         I-797 Approval & 33 & 33 & 100 \% \\
         I-797 Receipt & 71 & 69 & 97.18 \% \\
         \hline
    \end{tabular}
    \caption{Accuracy of supporting document classification.}
    \label{tbl:DocumentClassificationAccuracy}
\end{table}

\begin{figure}[ht]
    \centering
    \includegraphics[scale = 0.50]{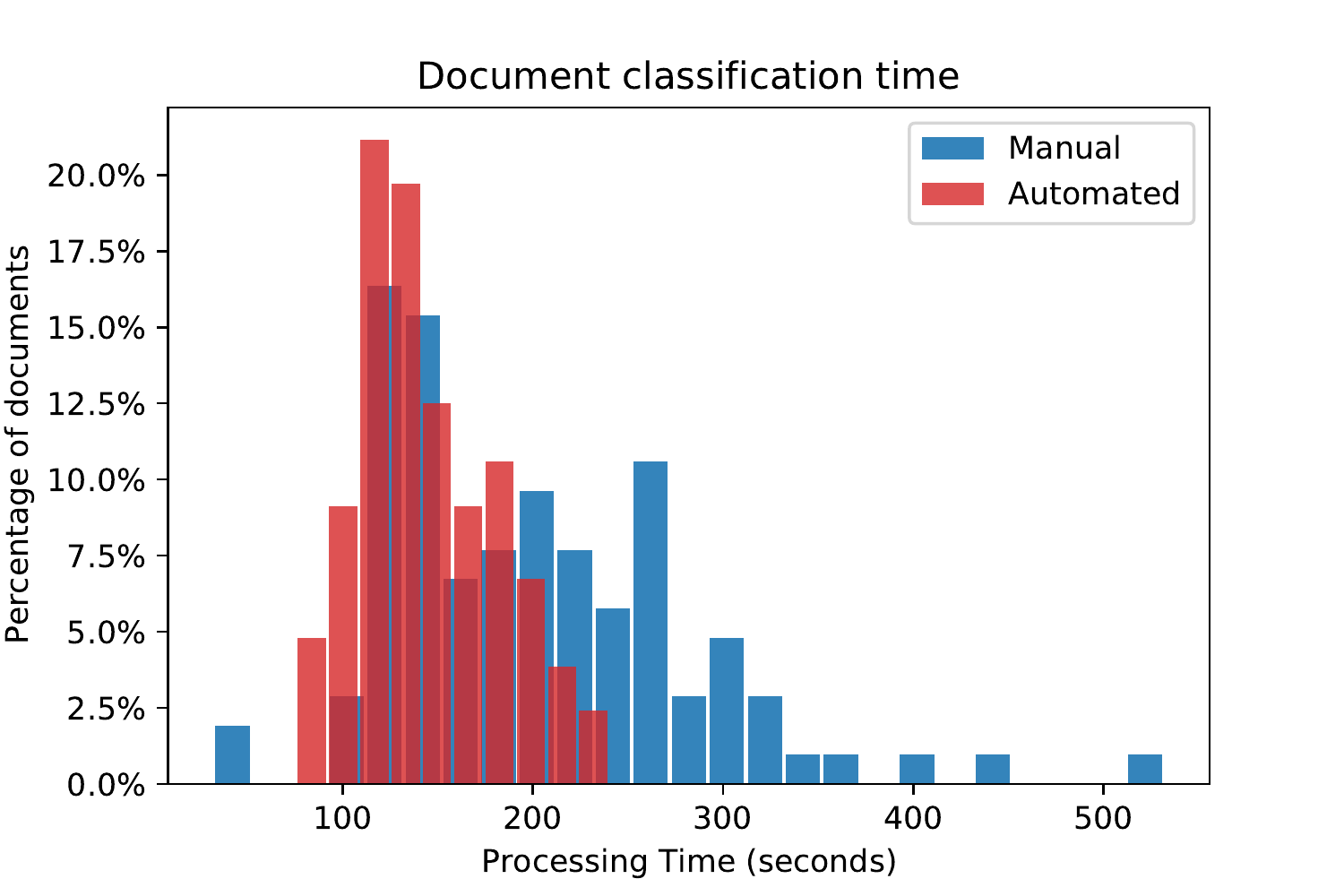}
    \caption{Histograms of document classification time (in seconds) using manual and automated workflows.}
    \label{fig:DocumentClassificationTimeCombined}
\end{figure}

\paragraph{Processing Time.} Figure \ref{fig:DocumentClassificationTimeCombined} compares histograms of processing time (measured in seconds) using the manual and automated document classification workflows, while Table \ref{tbl:DocumentClassificationTimeComparison} compares their means, medians, standard deviations, minimum and maximum values.

\begin{table}[ht]
    \centering
    \begin{tabular}{r l l}
        \hline
         &  Manual & Automated \\
         \hline
    mean (seconds) & 196.87 & 144.98 \\
    median (seconds) & 187.5 & 135.0 \\
    standard deviation (seconds) & 79.66 & 36.37 \\
    min (seconds) & 104 & 75.6 \\
    max (seconds) & 532 & 240 \\
    \hline
    \end{tabular}
    \caption{Comparison between processing times of manual and automated document categorization workflows. }
    \label{tbl:DocumentClassificationTimeComparison}
\end{table}

\subsection{RFE Attack Type Classification and Response Generation}
\label{subsec:Evaluation:RFEAttackTypeClassificationAndResponseGeneration}

\subsubsection{Experimental Setup}

\begin{figure}[ht]
    \centering
    \includegraphics[scale = 0.35]{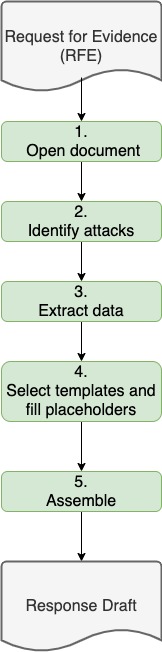}
    \caption{Experimental setup for Request For Evidence (RFE) attack type identification and response generation.}
    \label{fig:RFESetup}
\end{figure}

\begin{figure}[ht]
    \centering
    \includegraphics[scale = 0.45]{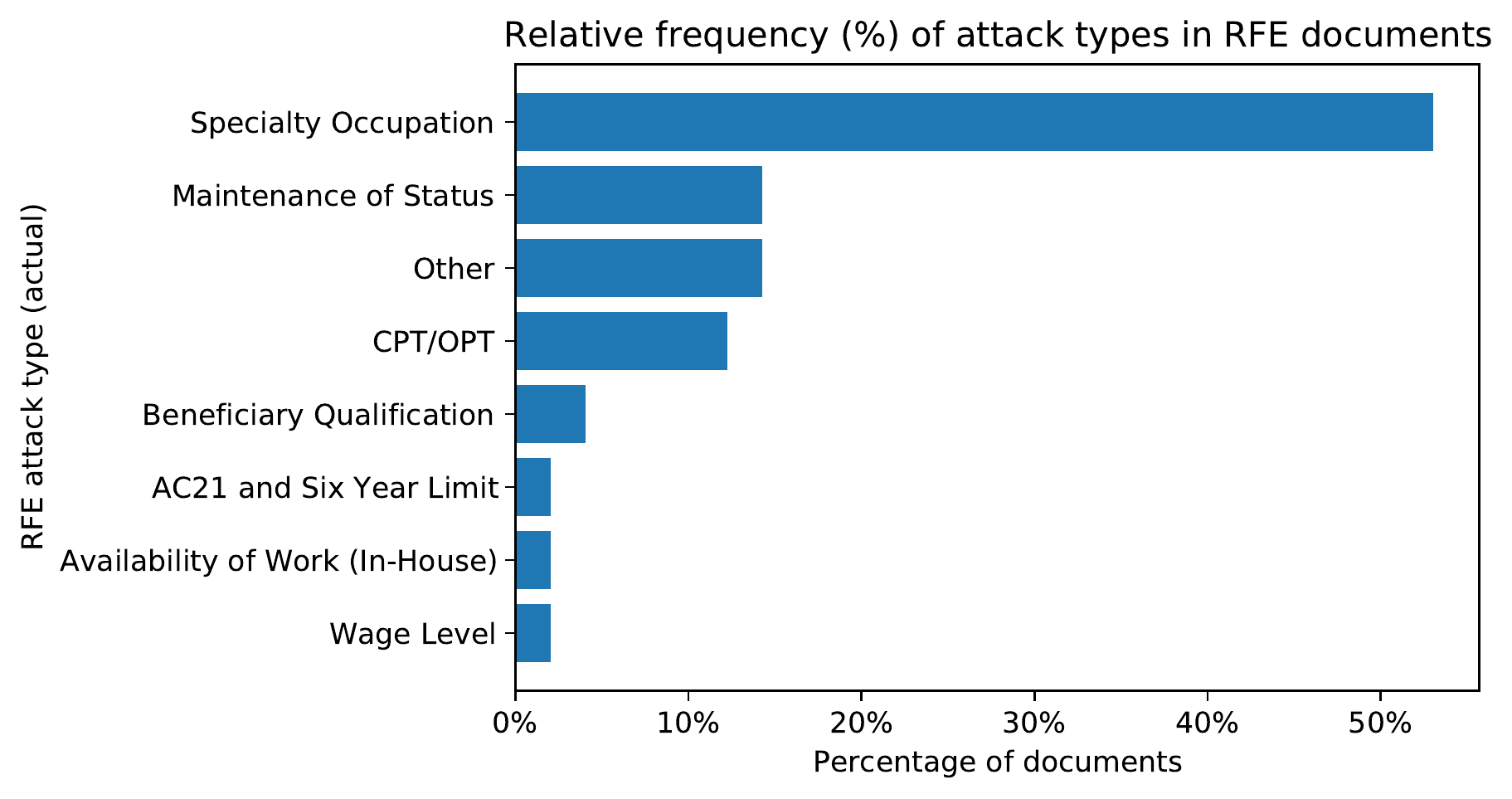}
    \caption{Frequency distribution of attack types in test set.}
    \label{fig:RFEFrequencyDistribution}
\end{figure}

Figure \ref{fig:RFESetup} depicts the workflow for responding to an RFE, consisting of opening the document, determining attack types based on textual content, extracting data from the RFE and beneficiary database, selecting response templates based on the attack types and extracted data, populating template placeholders with the values, and assembling the response. We compare the performance of manual and automated implementations of this workflow.
A test set of 49 RFEs containing various types of attacks was selected. The frequency distribution of attacks in this set is depicted in Figure \ref{fig:RFEFrequencyDistribution}. As seen in this figure, the predominant attack type is \textit{specialty occupation}, which is unsurprising since H-1B visa is only applicable to such occupations \cite{H1B}. Although our RFE attack classifier can classify a broad range of attacks, in this paper, we only focus on specialty occupation attacks.

\subsubsection{Results}

\paragraph{Accuracy.} 
Accuracy results for detection of specialty occupation attacks are presented in Table \ref{tbl:RFEAccuracy}. Here, a correct prediction refers to a scenario where an RFE contains a specialty occupation attack and the classifier flags the presence of this attack (true positive), or where an RFE does not contain a specialty occupation attack and the classifier does not flag this attack  (true negative). If an attack is present but undetected (false negative), or if an attack is flagged but not actually present (false positive), we consider the prediction incorrect. We define prediction accuracy as the fraction of predictions that are correct. We also measure the precision, recall, and F1-score\footnote{See, e.g., \url{https://scikit-learn.org/stable/auto_examples/model_selection/plot_precision_recall.html}}.
\begin{table}[ht]
    \centering
    \begin{tabular}{l r}
        \hline
        Metric & Value \\
        \hline
         Prediction accuracy (\%) & 73.47\% \\
         Precision &  0.7097\\
         Recall &  0.8462\\
         F1-score &  0.7719 \\
         \hline
    \end{tabular}
    \caption{Accuracy in detection of \textbf{specialty occupation} attacks in Request For Evidence (RFE) documents.}
    \label{tbl:RFEAccuracy}
\end{table}

\begin{figure}[ht]
    \centering
    \includegraphics[scale = 0.45]{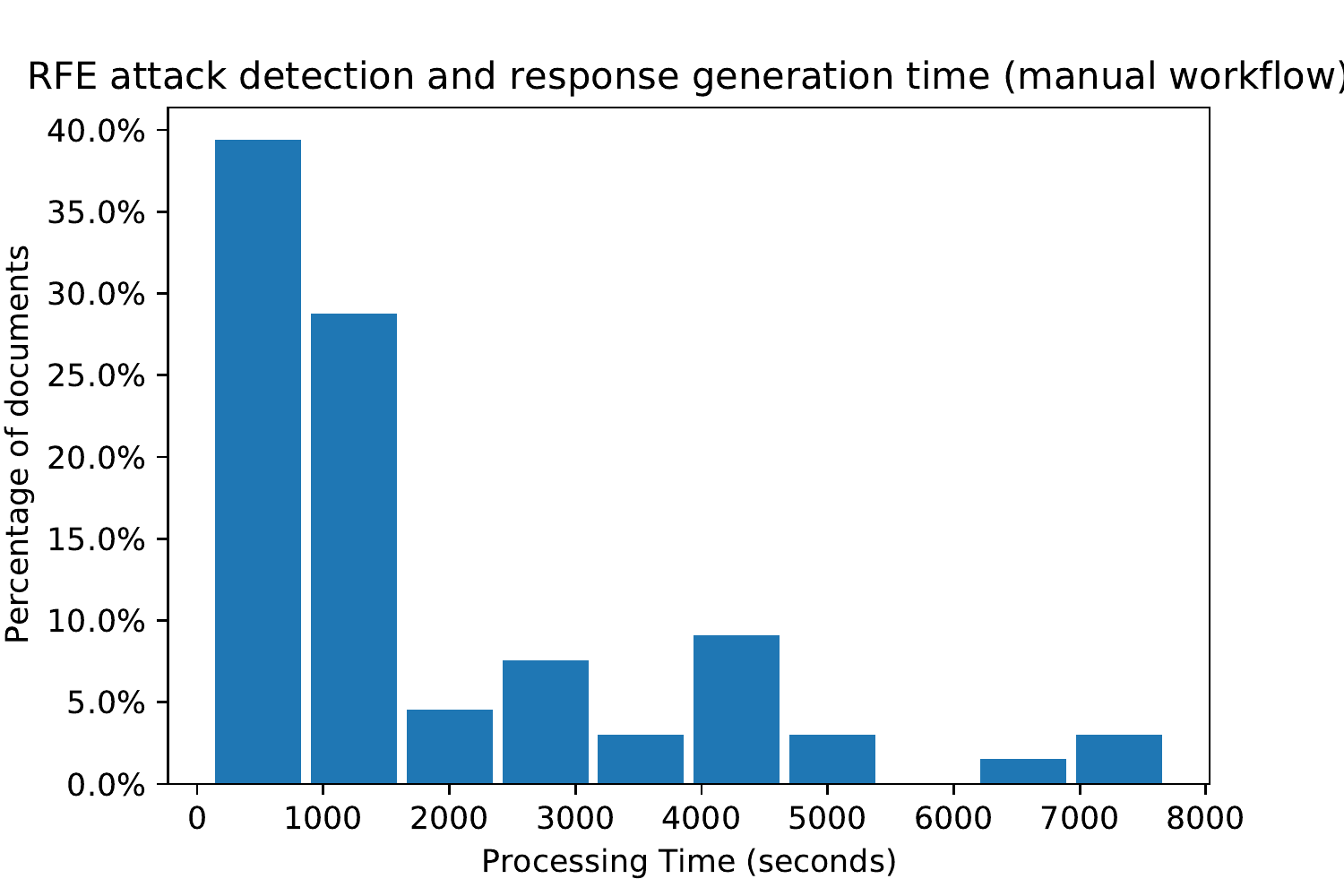} 
    \caption{Histogram of processing time (in seconds) for RFE attack type detection and response generation using a manual workflow.}
    \label{fig:RFEManual}
\end{figure}

\begin{figure}[ht]
    \centering
    \includegraphics[scale = 0.45]{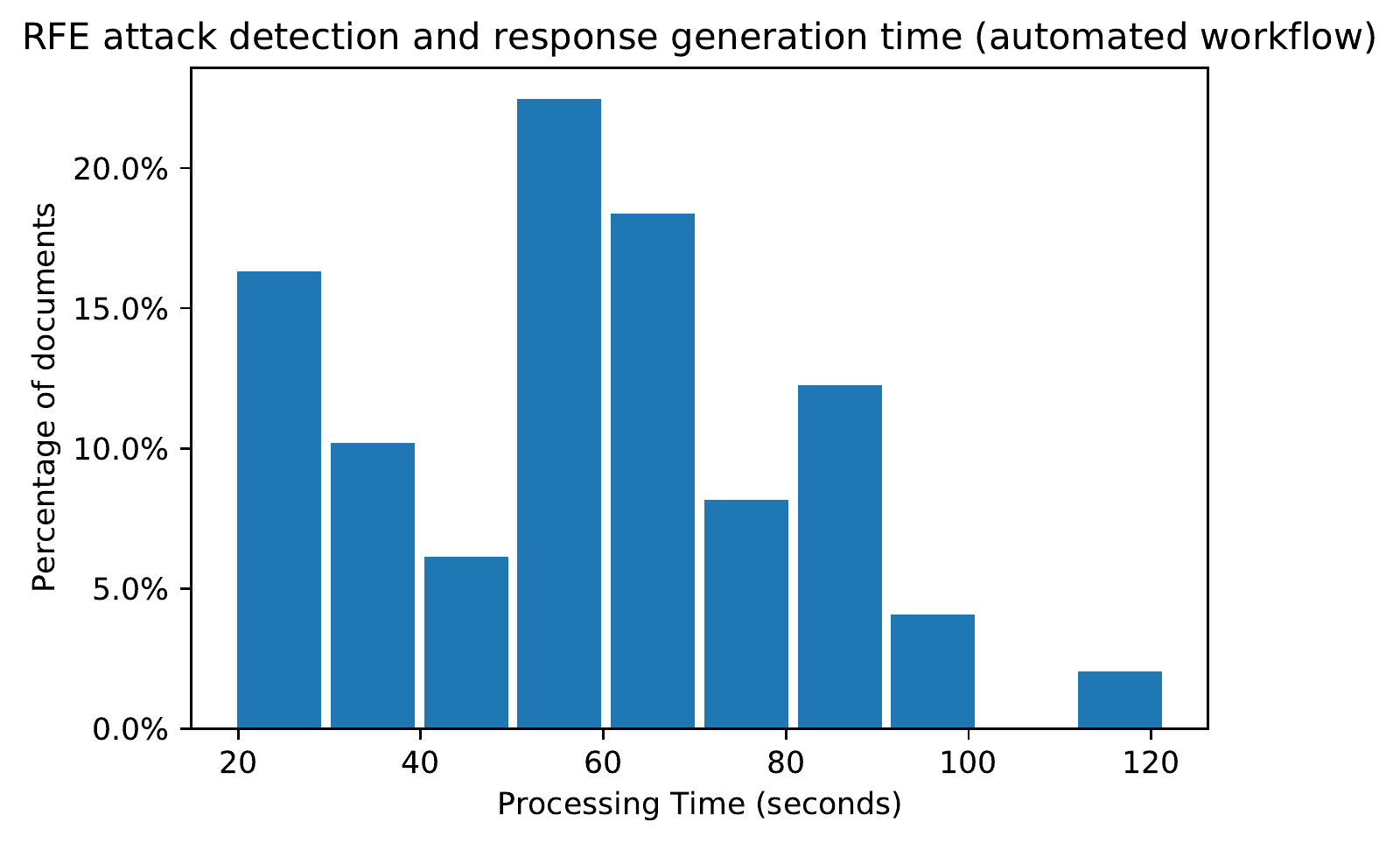} 
    \caption{Histogram of processing time (in seconds) for RFE attack type detection and response generation using our automated workflow.}
    \label{fig:RFEAutomated}
\end{figure}

\paragraph{Processing Time.} Figure \ref{fig:RFEManual} and Figure \ref{fig:RFEAutomated} show the histograms of processing time (measured in seconds) using manual and automated document classification workflows, respectively. We show these histograms in separate figures because the manual processing times are two orders of magnitude higher than the automated processing times, making it difficult to include them in the same figure.
Table \ref{tbl:RFEProcessingTime} compares their means, medians, standard deviations, minimum and maximum values.

\begin{table}[ht]
    \centering
    \begin{tabular}{r l l}
        \hline
         &  Manual & Automated \\
         \hline
    mean (seconds) & 1803.53  & 57.81 \\
    median (seconds) & 1189.0 & 57.99 \\
    standard deviation (seconds) & 1760.76 & 23.24 \\
    min (seconds) & 106 & 19.39 \\
    max (seconds) & 7692 & 121.69 \\
    \hline
    \end{tabular}
    \caption{Comparison between processing time (in seconds) of manual and automated workflows for RFE attack type identification and response generation. }
    \label{tbl:RFEProcessingTime}
\end{table}

In the next section, we interpret these results.

\section{Discussion}
\label{sec:Discussion}

The results presented in the previous section allow us to quantify the advantage of our approach over purely manual processes. In the supporting document classification problem, the classifier can distinguish documents of two different types with strong visual and textual similarities with an accuracy of 98.08\%. On the other hand, Table \ref{tbl:DocumentClassificationTimeComparison} shows that the automated workflow lowers the average document processing time from 196.87 seconds to 144.98 seconds, resulting in a reduction of 26.36 \%. Moreover, we also see that the standard deviation decreases from 79.66 seconds to 36.37 seconds, suggesting that the processing time of the automated workflow is more consistent. This is supported by Figure \ref{fig:DocumentClassificationTimeCombined} which shows that manual processing time has a much wider spread than automated processing time.

In the RFE attack type detection problem, Table \ref{tbl:RFEAccuracy} shows that the prediction accuracy is 73.47\%. We also note that the precision is 0.7097 and recall is 0.8462. In other words, out of all the RFE documents in which the classifier flags a specialty occupation attack, 70.97\% actually contain the attack. On the other hand, of all the RFE documents that actually contain the specialty occupation attack, our classifier is able to flag its presence in 84.62\% of the cases. In terms of processing time, the average end to end time decreases from 1803.53 seconds (approximately, 30 minutes) to 57.81 seconds (approximately, 1 minute), resulting in a reduction of 96.79\%.

In view of the above, our approach reduces the amount of repetitive manual work necessary in both problems. However, a human in the loop is necessary for reviewing the outputs of our automated workflows.

\section{Conclusion}
\label{sec:Conclusion}

In this paper, we have considered the problem of categorizing documents necessary for supporting U.S.~work visa petitions, as well as preparing responses to Requests For Evidence (RFE) issued by the U.S.~Citizenship and Immigration Services (USCIS). Typically, both processes are entirely manual, and a significant portion of the work is repetitive without requiring any human ingenuity. To reduce the burden of manual repetitive work,  we have demonstrated that machine learning methods may be applied to partially automate these workflows. In particular, we have used an ensemble of an image classifier and a text classifier to categorize supporting documents. We have also used a text classifier to detect the types of evidence being requested in an RFE, and used the identified types in conjunction with response templates and extracted fields to assemble draft responses. Finally, we have empirically demonstrated that these automated workflows achieve considerable accuracy while significantly reducing processing time.

\bibliographystyle{IEEEtran}
\bibliography{mybib.bib}

\begin{thebibliography}{10}
\providecommand{\url}[1]{#1}
\csname url@samestyle\endcsname
\providecommand{\newblock}{\relax}
\providecommand{\bibinfo}[2]{#2}
\providecommand{\BIBentrySTDinterwordspacing}{\spaceskip=0pt\relax}
\providecommand{\BIBentryALTinterwordstretchfactor}{4}
\providecommand{\BIBentryALTinterwordspacing}{\spaceskip=\fontdimen2\font plus
\BIBentryALTinterwordstretchfactor\fontdimen3\font minus
  \fontdimen4\font\relax}
\providecommand{\BIBforeignlanguage}[2]{{%
\expandafter\ifx\csname l@#1\endcsname\relax
\typeout{** WARNING: IEEEtran.bst: No hyphenation pattern has been}%
\typeout{** loaded for the language `#1'. Using the pattern for}%
\typeout{** the default language instead.}%
\else
\language=\csname l@#1\endcsname
\fi
#2}}
\providecommand{\BIBdecl}{\relax}
\BIBdecl

\bibitem{I797}
{U.S.~Citizenship and Immigration Services}, ``{Form I-797: Types and
  Functions},''
  \url{https://www.uscis.gov/forms/filing-guidance/form-i-797-types-and-functions},
  accessed: August 13, 2020.

\bibitem{I797C}
------, ``{Form I-797C, Notice of Action},''
  \url{https://www.uscis.gov/forms/form-i-797c-notice-of-action}, accessed:
  August 13, 2020.

\bibitem{EAD}
------, ``Employment authorization document,''
  \url{https://www.uscis.gov/green-card/green-card-processes-and-procedures/employment-authorization-document},
  accessed: August 13, 2020.

\bibitem{RFE}
------, ``Understanding requests for evidence (rfes): A breakdown of why rfes
  were issued for h-1b petitions in fiscal year 2018,''
  \url{https://www.uscis.gov/sites/default/files/document/data/understanding-requests-for-evidence-h-1b-petitions-in-fiscal-year-2018.pdf},
  accessed: August 14, 2020.

\bibitem{SOC}
{U.S. Bureau of Labor Statistics}, ``{2018 SOC Definitions},''
  \url{https://www.bls.gov/soc/2018/soc_2018_definitions.pdf}, accessed: August
  14, 2020.

\bibitem{Surden2014}
\BIBentryALTinterwordspacing
H.~Surden, ``Machine learning and law,'' \emph{89 WASH. L. REV.87}, 2014.
  [Online]. Available: \url{https://scholar.law.colorado.edu/articles/81}
\BIBentrySTDinterwordspacing

\bibitem{10.1007/978-3-030-19823-7_31}
N.~Bansal, A.~Sharma, and R.~K. Singh, ``A review on the application of deep
  learning in legal domain,'' in \emph{Artificial Intelligence Applications and
  Innovations}, J.~MacIntyre, I.~Maglogiannis, L.~Iliadis, and E.~Pimenidis,
  Eds.\hskip 1em plus 0.5em minus 0.4em\relax Cham: Springer International
  Publishing, 2019, pp. 374--381.

\bibitem{Faggella2020}
D.~Faggella, ``Ai in law and legal practice – a comprehensive view of 35
  current applications,''
  \url{https://emerj.com/ai-sector-overviews/ai-in-law-legal-practice-current-applications/},
  accessed: August 13, 2020.

\bibitem{10.2307/4099370}
\BIBentryALTinterwordspacing
T.~W. Ruger, P.~T. Kim, A.~D. Martin, and K.~M. Quinn, ``The supreme court
  forecasting project: Legal and political science approaches to predicting
  supreme court decisionmaking,'' \emph{Columbia Law Review}, vol. 104, no.~4,
  pp. 1150--1210, 2004. [Online]. Available:
  \url{http://www.jstor.org/stable/4099370}
\BIBentrySTDinterwordspacing

\bibitem{10.2307/3688543}
\BIBentryALTinterwordspacing
A.~D. Martin, K.~M. Quinn, T.~W. Ruger, and P.~T. Kim, ``Competing approaches
  to predicting supreme court decision making,'' \emph{Perspectives on
  Politics}, vol.~2, no.~4, pp. 761--767, 2004. [Online]. Available:
  \url{http://www.jstor.org/stable/3688543}
\BIBentrySTDinterwordspacing

\bibitem{Katz2016}
D.~Katz, I.~Bommarito, and J.~Blackman, ``A general approach for predicting the
  behavior of the supreme court of the united states,'' \emph{PLOS ONE},
  vol.~12, 12 2016.

\bibitem{Aletras2016PredictingJD}
N.~Aletras, D.~Tsarapatsanis, D.~Preotiuc-Pietro, and V.~Lampos, ``Predicting
  judicial decisions of the european court of human rights: a natural language
  processing perspective,'' \emph{PeerJ Comput. Sci.}, vol.~2, p. e93, 2016.

\bibitem{Medvedeva2019UsingML}
M.~Medvedeva, M.~Vols, and M.~Wieling, ``Using machine learning to predict
  decisions of the european court of human rights,'' \emph{Artificial
  Intelligence and Law}, vol.~28, pp. 237--266, 2019.

\bibitem{DBLP:conf/icail/YangGFY17}
\BIBentryALTinterwordspacing
E.~Yang, D.~A. Grossman, O.~Frieder, and R.~Yurchak, ``Effectiveness results
  for popular e-discovery algorithms,'' in \emph{Proceedings of the 16th
  edition of the International Conference on Artificial Intelligence and Law,
  {ICAIL} 2017, London, United Kingdom, June 12-16, 2017}, J.~Keppens and
  G.~Governatori, Eds.\hskip 1em plus 0.5em minus 0.4em\relax {ACM}, 2017, pp.
  261--264. [Online]. Available: \url{https://doi.org/10.1145/3086512.3086540}
\BIBentrySTDinterwordspacing

\bibitem{Cormack2014EvaluationOM}
G.~V. Cormack and M.~R. Grossman, ``Evaluation of machine-learning protocols
  for technology-assisted review in electronic discovery,'' \emph{Proceedings
  of the 37th international ACM SIGIR conference on Research \& development in
  information retrieval}, 2014.

\bibitem{Lemley2007}
M.~A. Lemley and J.~H. Walker, ``Intellectual property litigation
  clearinghouse: Data overview,'' \emph{Kauffman Symposium on Entrepreneurship
  and Innovation Data}, 2007.

\bibitem{DBLP:conf/bigdataconf/WeiQYZ18}
\BIBentryALTinterwordspacing
F.~Wei, H.~Qin, S.~Ye, and H.~Zhao, ``Empirical study of deep learning for text
  classification in legal document review,'' in \emph{{IEEE} International
  Conference on Big Data, Big Data 2018, Seattle, WA, USA, December 10-13,
  2018}, N.~Abe, H.~Liu, C.~Pu, X.~Hu, N.~K. Ahmed, M.~Qiao, Y.~Song,
  D.~Kossmann, B.~Liu, K.~Lee, J.~Tang, J.~He, and J.~S. Saltz, Eds.\hskip 1em
  plus 0.5em minus 0.4em\relax {IEEE}, 2018, pp. 3317--3320. [Online].
  Available: \url{https://doi.org/10.1109/BigData.2018.8622157}
\BIBentrySTDinterwordspacing

\bibitem{Silva2018DocumentTC}
N.~C. Silva, F.~Braz, T.~E. de~Campos, A.~B.~S. Guedes, D.~B. Mendes, D.~A.
  Bezerra, D.~B. Gusmao, F.~B.~S. Chaves, G.~G. Ziegler, L.~H. Horinouchi,
  M.~U. Ferreira, P.~H. Inazawa, V.~H.~D. Coelho, R.~V.~C. Fernandes, F.~H.
  Peixoto, M.~S.~M. Filho, B.~P. Sukiennik, L.~Rosa, R.~Silva, T.~A. Junquilho,
  and G.~Carvalho, ``Document type classification for brazil’s supreme court
  using a convolutional neural network,'' in \emph{ICoFCS-2018}, 2018.

\bibitem{DBLP:conf/fedcsis/UndaviaMO18}
\BIBentryALTinterwordspacing
S.~Undavia, A.~Meyers, and J.~Ortega, ``A comparative study of classifying
  legal documents with neural networks,'' in \emph{Proceedings of the 2018
  Federated Conference on Computer Science and Information Systems, FedCSIS
  2018, Pozna{\'{n}}, Poland, September 9-12, 2018}, ser. Annals of Computer
  Science and Information Systems, M.~Ganzha, L.~A. Maciaszek, and
  M.~Paprzycki, Eds., vol.~15, 2018, pp. 515--522. [Online]. Available:
  \url{https://doi.org/10.15439/2018F227}
\BIBentrySTDinterwordspacing

\bibitem{DBLP:conf/cikm/LuCAK11}
\BIBentryALTinterwordspacing
Q.~Lu, J.~G. Conrad, K.~Al{-}Kofahi, and W.~Keenan, ``Legal document clustering
  with built-in topic segmentation,'' in \emph{Proceedings of the 20th {ACM}
  Conference on Information and Knowledge Management, {CIKM} 2011, Glasgow,
  United Kingdom, October 24-28, 2011}, C.~Macdonald, I.~Ounis, and I.~Ruthven,
  Eds.\hskip 1em plus 0.5em minus 0.4em\relax {ACM}, 2011, pp. 383--392.
  [Online]. Available: \url{https://doi.org/10.1145/2063576.2063636}
\BIBentrySTDinterwordspacing

\bibitem{DBLP:conf/propor/FurquimL12}
\BIBentryALTinterwordspacing
L.~O. de~Colla~Furquim and V.~L.~S. de~Lima, ``Clustering and categorization of
  brazilian portuguese legal documents,'' in \emph{Computational Processing of
  the Portuguese Language - 10th International Conference, {PROPOR} 2012,
  Coimbra, Portugal, April 17-20, 2012. Proceedings}, ser. Lecture Notes in
  Computer Science, H.~de~Medeiros~Caseli, A.~Villavicencio, A.~J.~S. Teixeira,
  and F.~Perdig{\~{a}}o, Eds., vol. 7243.\hskip 1em plus 0.5em minus
  0.4em\relax Springer, 2012, pp. 272--283. [Online]. Available:
  \url{https://doi.org/10.1007/978-3-642-28885-2\_31}
\BIBentrySTDinterwordspacing

\bibitem{Kumar2012}
R.~K. V and K.~Raghuveer, ``Article: Legal documents clustering using latent
  dirichlet allocation,'' \emph{International Journal of Applied Information
  Systems}, vol.~2, no.~6, pp. 27--33, May 2012, published by Foundation of
  Computer Science, New York, USA.

\bibitem{DBLP:conf/afips/SprowlBCEK84}
\BIBentryALTinterwordspacing
J.~Sprowl, P.~Balasubramanian, T.~Chinwalla, M.~W. Evens, and H.~Klawans, ``An
  expert system for drafting legal documents,'' in \emph{American Federation of
  Information Processing Societies: 1984 National Computer Conference, 9-12
  July 1984, Las Vegas, Nevada, {USA}}, ser. {AFIPS} Conference Proceedings,
  vol.~53.\hskip 1em plus 0.5em minus 0.4em\relax {AFIPS} Press, 1984, pp.
  667--673. [Online]. Available: \url{https://doi.org/10.1145/1499310.1499396}
\BIBentrySTDinterwordspacing

\bibitem{Betts2017}
K.~D. Betts and K.~R. Jaep, ``The dawn of fully automated contract drafting:
  Machine learning breathes new life into a decades-old promise,'' \emph{15
  Duke Law \& Technology Review}, pp. 216--233, 2017.

\bibitem{Miller}
S.~Miller, ``Benefits of artificial intelligence: what have you done for me
  lately?''
  \url{https://legal.thomsonreuters.com/en/insights/articles/benefits-of-artificial-intelligence},
  accessed: August 16, 2020.

\bibitem{DBLP:conf/nips/YosinskiCBL14}
\BIBentryALTinterwordspacing
J.~Yosinski, J.~Clune, Y.~Bengio, and H.~Lipson, ``How transferable are
  features in deep neural networks?'' in \emph{Advances in Neural Information
  Processing Systems 27: Annual Conference on Neural Information Processing
  Systems 2014, December 8-13 2014, Montreal, Quebec, Canada}, 2014, pp.
  3320--3328. [Online]. Available:
  \url{http://papers.nips.cc/paper/5347-how-transferable-are-features-in-deep-neural-networks}
\BIBentrySTDinterwordspacing

\bibitem{DBLP:journals/corr/SimonyanZ14a}
\BIBentryALTinterwordspacing
K.~Simonyan and A.~Zisserman, ``Very deep convolutional networks for
  large-scale image recognition,'' in \emph{3rd International Conference on
  Learning Representations, {ICLR} 2015, San Diego, CA, USA, May 7-9, 2015,
  Conference Track Proceedings}, Y.~Bengio and Y.~LeCun, Eds., 2015. [Online].
  Available: \url{http://arxiv.org/abs/1409.1556}
\BIBentrySTDinterwordspacing

\bibitem{DBLP:conf/cvpr/DengDSLL009}
\BIBentryALTinterwordspacing
J.~Deng, W.~Dong, R.~Socher, L.~Li, K.~Li, and F.~Li, ``Imagenet: {A}
  large-scale hierarchical image database,'' in \emph{2009 {IEEE} Computer
  Society Conference on Computer Vision and Pattern Recognition {(CVPR} 2009),
  20-25 June 2009, Miami, Florida, {USA}}.\hskip 1em plus 0.5em minus
  0.4em\relax {IEEE} Computer Society, 2009, pp. 248--255. [Online]. Available:
  \url{https://doi.org/10.1109/CVPR.2009.5206848}
\BIBentrySTDinterwordspacing

\bibitem{DBLP:conf/ecml/Joachims98}
\BIBentryALTinterwordspacing
T.~Joachims, ``Text categorization with support vector machines: Learning with
  many relevant features,'' in \emph{Machine Learning: ECML-98, 10th European
  Conference on Machine Learning, Chemnitz, Germany, April 21-23, 1998,
  Proceedings}, ser. Lecture Notes in Computer Science, C.~Nedellec and
  C.~Rouveirol, Eds., vol. 1398.\hskip 1em plus 0.5em minus 0.4em\relax
  Springer, 1998, pp. 137--142. [Online]. Available:
  \url{https://doi.org/10.1007/BFb0026683}
\BIBentrySTDinterwordspacing

\bibitem{10.5555/1288165.1288167}
A.~Kay, ``Tesseract: An open-source optical character recognition engine,''
  \emph{Linux J.}, vol. 2007, no. 159, p.~2, Jul. 2007.

\bibitem{scikit-learn}
F.~Pedregosa, G.~Varoquaux, A.~Gramfort, V.~Michel, B.~Thirion, O.~Grisel,
  M.~Blondel, P.~Prettenhofer, R.~Weiss, V.~Dubourg, J.~Vanderplas, A.~Passos,
  D.~Cournapeau, M.~Brucher, M.~Perrot, and E.~Duchesnay, ``Scikit-learn:
  Machine learning in {P}ython,'' \emph{Journal of Machine Learning Research},
  vol.~12, pp. 2825--2830, 2011.

\bibitem{chollet2015keras}
F.~Chollet \emph{et~al.}, ``Keras,'' \url{https://keras.io}, 2015.

\bibitem{abadi2016tensorflow}
M.~Abadi, P.~Barham, J.~Chen, Z.~Chen, A.~Davis, J.~Dean, M.~Devin,
  S.~Ghemawat, G.~Irving, M.~Isard \emph{et~al.}, ``Tensorflow: A system for
  large-scale machine learning,'' in \emph{12th $\{$USENIX$\}$ Symposium on
  Operating Systems Design and Implementation ($\{$OSDI$\}$ 16)}, 2016, pp.
  265--283.

\bibitem{H1B}
{U.S.~Citizenship and Immigration Services}, ``6.5 {H-1B} specialty
  occupations,''
  \url{https://www.uscis.gov/i-9-central/form-i-9-resources/handbook-for-employers-m-274/60-evidence-of-status-for-certain-categories/65-h-1b-specialty-occupations},
  accessed: August 19, 2020.

\end{thebibliography}

\end{document}